# Reducing the Model Variance of a Rectal Cancer Segmentation Network

**Joohyung Lee[1], Ji Eun Oh[1], Min Ju Kim[2], Bo Yun Hur[2], and Dae Kyung Sohn[1,2]**

[1]Innovative Medical Engineering & Technology, National Cancer Center Research Institute, Goyang-si, Gyeonggi-do 10408 South Korea
[2]Center for Colorectal Cancer, National Cancer Center, Goyang-si, Gyeonggi-do 10408 South Korea

Corresponding author: Dae Kyung Sohn (e-mail: gsgsbal@ncc.re.kr).

This project was supported by a grant from the National Cancer Center (Grant Number NCC-1710070)

**ABSTRACT** In preoperative imaging, the demarcation of rectal cancer with magnetic resonance images provides an important basis for cancer staging and treatment planning. Recently, deep learning has greatly improved the state-of-the-art method in automatic segmentation. However, limitations in data availability in the medical field can cause large variance and consequent overfitting to medical image segmentation networks. In this study, we propose methods to reduce the model variance of a rectal cancer segmentation network by adding a rectum segmentation task and performing data augmentation; the geometric correlation between the rectum and rectal cancer motivated the former approach. Moreover, we propose a method to perform a bias-variance analysis within an arbitrary region-of-interest (ROI) of a segmentation network, which we applied to assess the efficacy of our approaches in reducing model variance. As a result, adding a rectum segmentation task reduced the model variance of the rectal cancer segmentation network within tumor regions by a factor of 0.90; data augmentation further reduced the variance by a factor of 0.89. These approaches also reduced the training duration by a factor of 0.96 and a further factor of 0.78, respectively. Our approaches will improve the quality of rectal cancer staging by increasing the accuracy of its automatic demarcation and by providing rectum boundary information since rectal cancer staging requires the demarcation of both rectum and rectal cancer. Besides such clinical benefits, our method also enables segmentation networks to be assessed with bias-variance analysis within an arbitrary ROI, such as a cancerous region.

## I. INTRODUCTION

Globally, colorectal cancer is the third most common cancer and the second leading cause of cancer mortality [1]. Specifically, colorectal cancer was the most commonly diagnosed cancer in Korea in 2017, with 27,837 new cases [2]. The global burden of colorectal cancer is rising rapidly and is expected to increase by 60% by 2030 [3].

Depending on the cancer site, colorectal cancer can be defined as colon cancer or rectal cancer [1]. The Union for International Cancer Control's TNM Classification of Malignant Tumors (8th edition) categorizes rectal cancer as a tumor starting in the rectum, i.e., the last 12 centimeters of the colon [4]. The T-categorization of rectal cancer, a widely used rectal cancer staging criterion, pathologically classifies its progression by the degree of tumor invasion into the rectal wall. In magnetic resonance (MR) images, the T-category is determined by the relative location of rectal cancer and the rectal wall [5]. Since current treatment guidelines for rectal cancer utilize the T-category to recommend clinical treatments, accurate segmentation of rectal cancer is crucial. However, in practice, radiologists manually locate rectal cancer using MR images. Manual localization is time-consuming, and a reliable automatic segmentation system is necessary [6].

In recent years, deep learning has improved the state-of-the-art methods in various fields related to computer vision [7]. Its wide applicability derives from its ability to find complex structures in high-dimensional data. The introduction of convolutional neural networks (CNNs) has enhanced the ability of deep learning to learn a complex representation of images. Its performance has been further improved by the incorporation of new network backbones and convolution block [8-12].

Deep learning has also proved its applicability in various medical image analysis tasks, including medical image segmentation [13]. For example, Ronneberger *et al.* have introduced the U-Net by implementing a VGG-Net-like encoder together with a mirrored decoder for cell segmentation [14]. This encoder-decoder approach implements a fully convolutional network which is known to improve the computational efficiency of patch-based segmentation methods [6, 15]. Further, Milletari *et al.* have extended the U-Net to 3D images and introduced the Dice Similarity Coefficient (DSC)-based loss function for the segmentation of the prostate volume in magnetic resonance imaging (MRI) [16].

However, automation of medical image analysis remains challenging due to the inherent complexity of medical images and the extensive variation between patients [17]. Such complexity and large variability within data call for a model with a large capacity such as a deep neural network (DNN), able to discover intricate structures in the data. However, since high-capacity models can fit the intricate details of the data, they are usually less robust to data variation, and prone to overfitting, unless trained with many samples [18]. Unfortunately, in practice, relatively few annotated images are available in the medical field, so that overfitting can be a problem in building DNN models for medical image analysis.

There have been various attempts to moderate overfitting in deep learning, such as batch normalization, drop-out, data augmentation, image normalization, etc. [19-21]. In addition, multi-task learning (MTL) is known to reduce the risk of overfitting [18, 22]. By adding a different task, the parameters of the model are optimized towards values that can explain the variation observed in both tasks, thus reducing the risk of overfitting for the original network. In the case of a DNN model, the shared portion of the MTL network can be constrained towards values with better generalization ability if the additional task provides information relevant to the original task. Therefore, adding another task will reduce the risk of overfitting, if the additional task is related to the original one.

The risk of overfitting can be assessed by bias-variance analysis since overfitting is caused by high variance [18]. Specifically, a bias-variance analysis decomposes the generalization error into model bias and model variance. The analysis evaluates the model variance by creating multiple models from a single learner by varying the learner training sets. By varying the training set, bias-variance analysis can assess if the model is robust to data variation. If the learner is not robust and cannot generalize the data well, varying the training set will cause highly variant models. As a result, the risk of overfitting can be reduced by lowering the model variance.

Although such model robustness can also be measured by the discrepancy between training accuracy and validation accuracy, selection bias in choosing the training and validation sets can be a problem. In fact, selection bias can be critical especially for medical data, due to their limited size. Also, measuring the discrepancy between training accuracy and validation accuracy cannot capture the model robustness within a specific Region-of-Interest (ROI). However, in medical image analysis, model performance is especially important in the regions adjacent to the positive (cancerous) area. Consequently, a method which can measure the model robustness within an arbitrary ROI can help building models for medical image analysis.

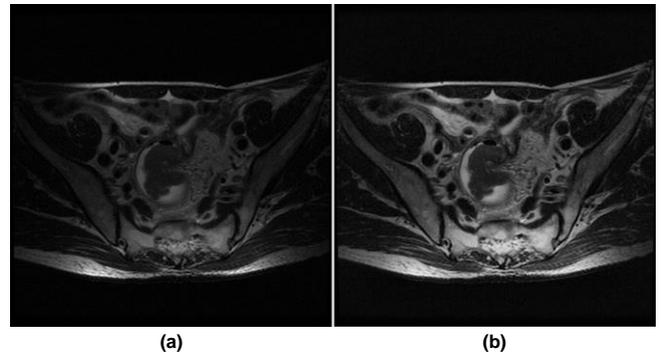

**FIGURE 1. Sample image before and after the preprocessing. (a) original image; (b) preprocessed image.**

Model variance is important to DNN-based medical image analysis models, but model bias cannot be ignored either. In fact, model bias contributes to the expected loss between ground truth and prediction through a trade-off relationship with model variance [18]. Although the mean squared loss is often used to derive the bias-variance decomposition, the unified theorem of bias-variance decomposition enables arbitrary loss functions to be decomposed into noise, bias, and variance, without loss of generality [23]. Despite the importance of model variance, we are not aware of any report that suggests a method to perform a bias-variance analysis of a segmentation model.

Also, we have not come across any study reporting an automatic system for segmenting both rectal cancer and rectum at once using MR images, although cancer staging indeed requires the demarcation of both rectum and rectal cancer. However, for the automatic segmentation of rectal cancer, Trebeschi *et al.* proposed a patch-based CNN model using both T2-weighted and diffusion-weighted images from MRI [6]. As a validation method, all image data were equally divided between the training and validation sets, which might lead to selection bias. Also, the post-processing method ran the risk of removing valid tumor areas, except for the largest tumor region.

In this study, we propose a pixel-wise bias-variance decomposition method to measure the model variance of segmentation network. This pixel-wise approach can not only measure the expected model bias and variance within an arbitrary ROI but can also visualize the bias and variance map of sample image. We also suggest two methods to reduce the model variance of rectal cancer segmentation network: 1) multi-region segmentation network (MRSN) by adding rectum segmentation task to rectal cancer segmentation network; and 2) the augmentation method that resizes each mini-batch into a random scale.

The efficacy of two proposed methods in reducing the model variance is validated in Section III A by the suggested pixel-wise bias-variance decomposition. Section II D will explain the proposed pixel-wise bias-variance decomposition in detail whereas Section II C and B describe MRSN and suggested augmentation method in detail, respectively. Note that in this study the term "model" denotes a learner whose parameters have been optimized using a training set, whereas "network" denotes a DNN learner.

## II. MATERIALS AND METHODS

### A. DATA PREPARATION

The experiment was performed using MR images of 1,813 rectal cancer patients, obtained between September 2004 and June 2016 at the National Cancer Center of South Korea. Among these cases, 457 were selected after disregarding the cases with at least one of the following properties: preoperative chemo-radiation, incomplete pathologic stage information, disagreement between MR image and pathologic staging, pathologic stage T1 or T4, tumors located more than 13 cm or less than 3 cm from the anal verge, or the application of either clipping or stents. The whole study was conducted according to the principles of the Declaration of Helsinki, and the protocol was approved by the Institutional Review Board of our institution (NCC2017-0031).

Rectal MRI examination was performed with one of four 3T or 1.5T superconducting systems: Achieva 3.0T (n = 233) and Achieva 3.0T TX (n = 131), by Philips Healthcare (Cleveland, OH, USA); or Signa HDX 3.0T (n = 19) and Genesis Signa 1.5T (n = 74), by GE Healthcare (Milwaukee, WI, USA), using pelvic phased-array coils. Among the various MRI sequences, our experiment evaluated axial T2-weighted fast-spin echo images.

Among approximately 30 image slices per patient, we selected one or two to create the dataset. The 907 selected images clearly reflected the T-category of the patient by showing the rectum with clear appearance of either T2 or T3 rectal cancer. Two gastrointestinal clinical specialists were involved not only in selecting the 907 image slices from the 457 cases, but also in the manual delineation of both rectum and rectal cancer. Specifically, one specialist drew the boundary, and the other specialist confirmed the outcome. These manual segmentation results were used as our ground truth.

For bias-variance analysis, we set apart 10% of the 907 images as a test set. Then, we used the remaining 90% to create nine training sets for which we performed 9-fold cross-validation. Nine-fold cross-validation was adopted only to create nine training sets, so we disregarded the nine validation sets thus created. With these nine different training sets, we created nine different models per network. Note that we did not create many training sets by a random sampling method such as bootstrapping since the training of many DNN models is time-consuming. Instead, all networks shared the same nine training sets and the single test set. This approach not only allowed for the fair comparison of learners by sharing the same nine training sets among different learners, but also allowed using all the available data efficiently.

### B. PREPROCESSING

We applied both image intensity range normalization and histogram equalization to improve image contrast and generalization [24]. As a normalization step, 90% of both the maximum and the minimum intensity value from the overall image slices of a patient were used to reduce the image depth from 12-bit to 8-bit. We also applied contrast-limited adaptive histogram equalization to enhance the contrast as well as to reduce the illumination effect [25-27]. As shown in Fig. 1, an image with a high-intensity artifact region, which decreases the overall image contrast, became more interpretable after preprocessing.

Motivated by Dao *et al.* [28], we also performed data augmentation to reduce the model variance of our proposed rectal cancer segmentation network, described in Section II C. Especially, we aimed to enhance the scale-robustness of our segmentation system since our raw MR images have heterogeneous scales (from 512×512 to 1056×1056) depending on the MR scanner and its settings. It should be noted that equalizing the pixel spacing of all images does not invalidate the need for scale-robustness; the anatomical structures in MR images can still differ in scale while being similar in shape even if the pixel spacing is equalized among all images. Above all, equalizing pixel spacing is infeasible because fixing the pixel spacing will make the size of images within a mini-batch heterogeneous. To enhance the scale-robustness of our segmentation network, we resized each mini-batch into a randomly chosen scale (ranging from 192×192 to 288×288); the fully convolutional nature of our network allows input images to have different sizes. Note that we did not crop images after random resizing to synchronize the size of all training images. Moreover, we did not create an image pyramid nor supplemented an additional network for scale-robustness, which would have increased the computational cost and the implementation complexity [29]. Instead, we trained the network with images at heterogeneous scales and with their original field-of-views uncropped. Given that medical images usually vary in scale due to the variability of the scanners and their settings, this augmentation method is expected, in general, to enhance the scale-robustness of other medical image analysis systems as well. The efficacy of this augmentation approach is evaluated in Sections III A and B.

Beside scale augmentation, we also performed the conventional random augmentation of the training images, i.e., adjusting the contrast, brightness, and sharpness, followed by a rotation, flipping (left and right), and cropping (maximum 10% from the edge and preserving the square shape). It should be noted that neither the validation nor the test data were augmented, but just resized to the single scale of 256×256.

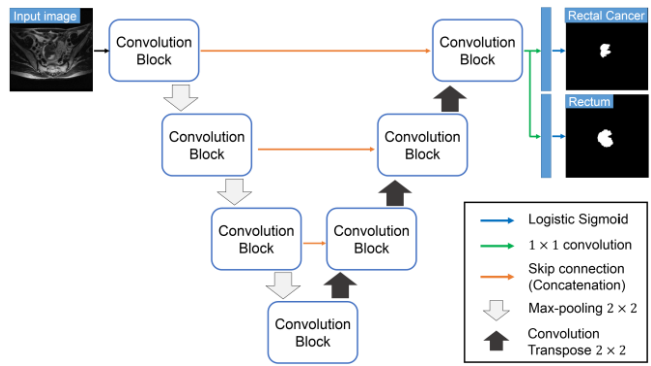

**FIGURE 2. Rectal cancer segmentation network with an additional task of rectum segmentation (MRSN).**

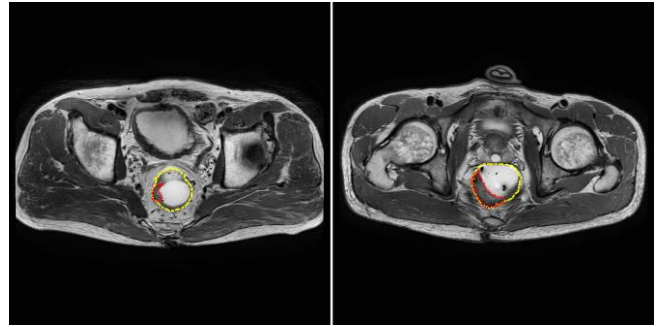

**FIGURE 3. Two different sample images with their ground truth overlaid. The red line indicates rectal cancer whereas the yellow line represents rectum. Note that the bladder in the left image appears similar to the rectum.**

## C. SEGMENTATION NETWORK ARCHITECTURE

We developed an encoder-decoder segmentation network to improve the computational efficiency of an existing rectal cancer segmentation method [6]. The first convolution layer of the network involves forty 3×3 filters. The number of filters per layer at the encoder is doubled after each convolution block as in the VGG-Net or U-Net neural networks [9, 14]. The decoder is a mirrored version of the encoder, and the number of filters per layer is halved through convolution transpose. Appendix A describes how the convolution block, which is illustrated in Fig. 6-(c), is selected for our network.

As depicted in Fig. 2, we added the rectum segmentation task to reduce the model variance of the rectal cancer segmentation network. The geometric correlation between the rectum and rectal cancer, which can be noticed from Fig. 3, motivated us to adopt this MTL-based approach. Specifically, rectal cancer is mostly located inside the rectum since it grows from the rectum area, which can be found in Fig. 3 [5]. Since our dataset only includes images that clearly reflect either T2- or T3-stage rectal cancer, rectal cancer is always found along the rectum wall [4]. Moreover, rectum and rectal cancer often share some portion of their boundaries, as can also be seen in Fig. 3. By adding a rectum segmentation task, our network yields the prediction for rectum boundary as a by-product, which is clinically informative as well, especially in cancer staging.

To implement the additional segmentation task, we added an additional task-specific 1×1 convolution layer for rectum segmentation after the last convolution block, as shown in Fig. 2. Note that we did not use a softmax function, but calculated the probability of both classes by logistic sigmoid after their own task-specific convolution layer, because rectal cancer and rectum can overlap each other and thus are not mutually exclusive. In this paper, single-region segmentation network (SRSN) denotes the network without additional task-specific layer, whereas multi-region segmentation network (MRSN) denotes the network with two parallel task-specific layers. In addition, MRSN-AUG denotes the MRSN with data augmentation based on image resizing, as described in Section II B. Consequently, the SRSN can segment only a single region, either rectum or rectal cancer, whereas both MRSN and MRSN-AUG segment both regions at once.

The efficacy of both MTL and image resizing-based augmentation in reducing model variance were evaluated by bias-variance analysis. The bias and variance of the SRSN, MRSN, and MRSN-AUG are compared in Section III A, whereas their segmentation performance based on DSC with 10-fold cross-validation are compared in Section III B.

All filter weights were initialized using the normal distribution sampling method suggested by He *et al.* [26], except for the transposed convolution filters, which were initialized using the uniform distribution sampling method suggested by Glorot and Bengio [30, 31]. The Adam optimization algorithm was implemented to stochastically optimize the parameters with a mini-batch size of 20 [32]. Due to the preponderance of negative pixels, we implemented the DSC-based loss function suggested by Milletari *et al.* as our optimization objective function, which can be written as

$$D = \frac{-2\sum_i^N p_i g_i}{\sum_i^N p_i^2 + \sum_i^N g_i^2} \qquad (1)$$

where the sums run over the N pixels, the predicted binary segmentation pixel being indicated by $p_i \in P$ and the ground truth binary pixel by $g_i$ [16].

## D. PIXEL-WISE BIAS-VARIANCE ANALYSIS FOR SEGMENTATION NETWORKS

We propose a method to quantify the bias, variance, and expected loss of a segmentation network within an arbitrary ROI, such as a cancerous region. This method allows us to confirm if both the additional rectum segmentation task and the augmentation method based on image resizing reduce the model variance of the rectal cancer segmentation network without increasing the model bias.

We measured bias and variance in accordance with the unified definition suggested by Domingos [23]. However,

two additional conditions should be considered for our problem. First, we have a single ground truth per test sample image. Second, our prediction is a multi-dimensional vector since the segmentation network predicts an image. Considering these two additional conditions, we decided to perform a pixel-wise bias-variance analysis. Then we calculated the expected values of bias, variance, and expected loss within an arbitrary area by averaging. Our approach for generating the training sets and the test set is illustrated in Section II A.

The main prediction for the test image $x$ at pixel $i$ for a loss function $L$ and a set of training sets $D$ becomes

$$y_m^{L,D}(x_i) = argmin_{y'} E_D[L(y(x_i), y')] \qquad (2)$$

where $y(x_i)$ is the prediction value for the test image $x$ at pixel $i$. We can specify our loss function as a zero-one loss since our problem is a pixel-wise classification problem. Considering that we have nine training sets, the main prediction at pixel $i$ becomes the mode among nine binary predictions at pixel $i$. Now, bias and variance can be defined using the main prediction.

The bias of a network for the test image $x$ at pixel $i$ is

$$B(x_i) = L(t(x_i), y_m(x_i)) \qquad (3)$$

where $t(x_i)$ represents the ground truth of the test image $x$ at pixel $i$. $y_m(x_i)$ is the main prediction which is the mode, as stated above. The variance of the test image $x$ at pixel $i$ can be defined as

$$V(x_i) = E_D[L(y_m(x_i), y(x_i))] \qquad (4)$$

With bias and variance thus defined, the expected loss of the image $x$ at pixel $i$ can be decomposed as

$$E_D[L(t(x_i), y(x_i))] = B(x_i) + c_2 V(x_i) \qquad (5)$$

where $c_2 = 1$ if $y_m(x_i) = t(x_i)$ (unbiased prediction), whereas $c_2 = -1$ if the prediction is biased [23]. Finally, we can measure the expected values of the bias, variance, and expected loss within an arbitrary ROI where $i \in I_{ROI}$ as follows:

$$E_{D,i}[L(t(x_i), y(x_i))] = E_i[B(x_i)] + E_i[c_2 V(x_i)] \qquad (6)$$

Now, we can compare the three different rectal cancer segmentation networks (i.e., SRSN, MRSN, and MRSN-AUG) in terms of their expected values of bias, variance, and zero-one loss within an arbitrary ROI. We measured the expected values over the entire image as well as over the positive (cancerous) region of the test images and reported the results in Section III A. We calculated the expected values within positive regions for two reasons. First, segmentation performance is more important in the positive than in the negative region. Second, the negative region contains an excess of non-body area, and segmentation models usually classify non-body regions correctly without difficulty. Consequently, including negative regions can excessively dilute the expected values (bias and variance); this makes it needlessly hard to prove a statistically significant difference between networks with these expected values. We performed statistical tests to objectively compare the distributions of the expected values per test sample from different networks.

## E. PERFORMANCE EVALUATION WITH CROSS-VALIDATION

Along with bias-variance analysis, we tested if our approach to reduce the model variance would also demonstrate improvement when evaluated by the conventional evaluation scheme. Specifically, we measured DSC, sensitivity, and specificity, via 10-fold cross-validation of 907 images, to compare the performance of SRSN, MRSN, and MRSN-AUG. We discuss the results in Section III B. Moreover, we used the same evaluation method to compare the performance of different convolution blocks, as described in Appendix A. DSC is a widely used metric in medical image segmentation tasks due to its robustness to highly imbalanced classes [24]. The DSC between two sets $A$ and $B$ (e.g., prediction and ground truth) is defined as

$$DSC(A, B) = \frac{2|A \cap B|}{|A| + |B|} \qquad (7)$$

## III. RESULTS AND DISCUSSION

### A. PIXEL-WISE BIAS-VARIANCE ANALYSIS

We compared the model variance of three different rectal cancer segmentation networks in order to assess the efficacy of our proposed methods, i.e., the addition of a rectum segmentation task and the augmentation method based on image resizing. Considering the geometric correlation of the rectum to rectal cancer as well as the heterogeneous scale of our MR images, we assumed that both our methods would reduce the variance of the rectal cancer segmentation network. To this end, the test set was predicted by nine different models trained on nine different training sets, as described in Section II A.

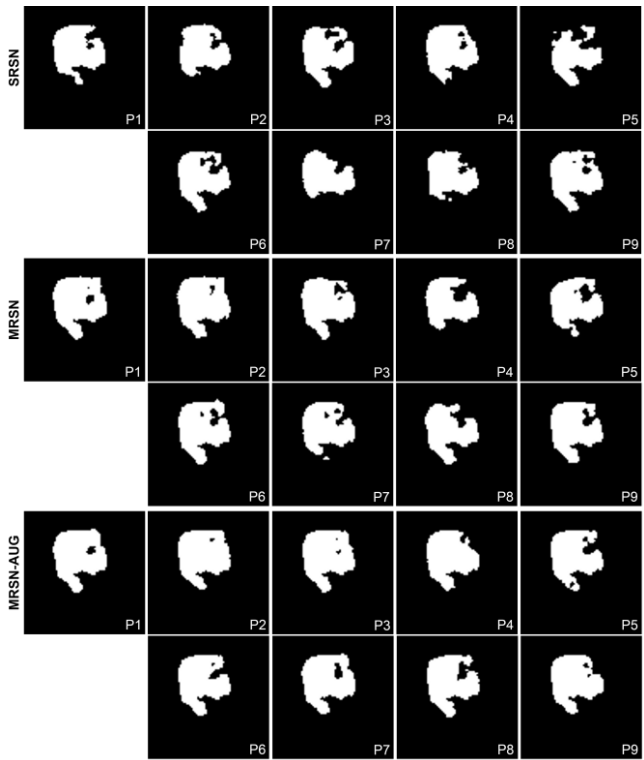

**FIGURE 4.** Nine rectal cancer prediction maps of a test image, generated by SRSN, MRSN, and MRSN-AUG. Each map is generated by one of three networks, optimized with one of the nine training sets, as described in Section II A. The SRSN, MRSN, and MRSN-AUG networks share training data only if their prediction numbers, located at the bottom right corner of each map with a "P" prefix, are the same. The ground truth of this example can be found in column (d) of Fig. 5 (upper example).

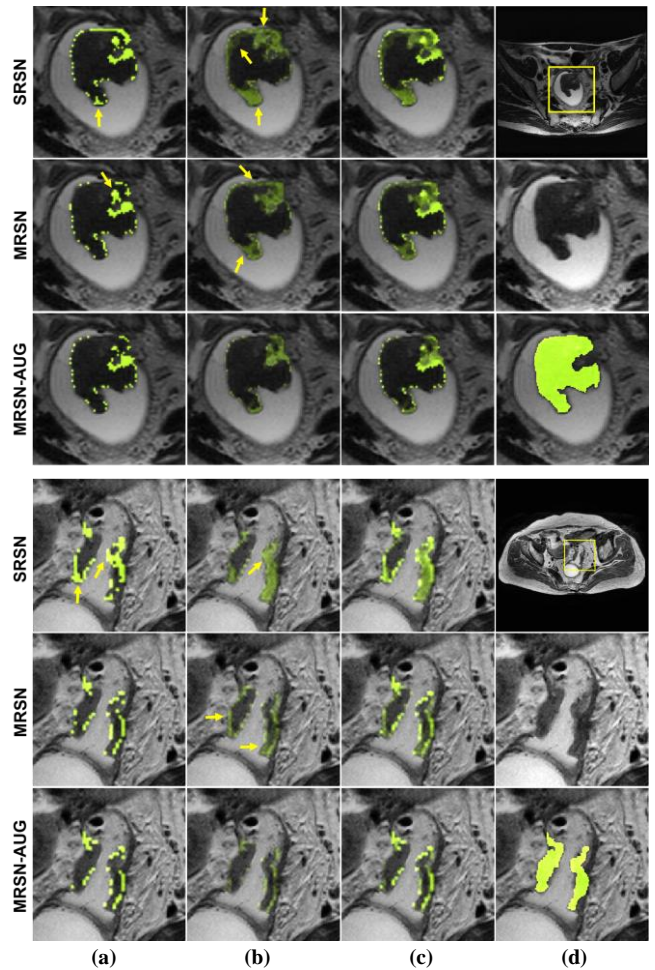

**FIGURE 5.** Two test images overlaid with their pixel-wise bias, variance, and expected loss maps produced by three different rectal cancer segmentation networks, i.e., SRSN, MRSN, and MRSN-AUG. Higher values are represented by colors tending more towards yellow than green. Column (d) shows the original images overlaid with the yellow cropping boundary, as well as the cropped images overlaid with the ground truth. The cropping boundary is set to visualize the rectal cancer region more closely. Note that the upper example corresponds to the one shown in Fig. 4.

Fig. 4 shows an example of the nine different rectal cancer prediction maps generated by the three different networks, i.e., SRSN, MRSN, and MRSN-AUG. The ground truth for this test sample is provided in Fig. 5, column (d). From each network, nine prediction maps (P1-P9) were generated by varying the training set. On the other hand, three prediction maps were generated from three different networks using each training set, namely prediction maps P1, P2, …, P9 for SRSN, MRSN, and MRSN-AUG. Large variations among the nine prediction maps indicate that the network cannot generalize well upon variation of the training data, which can lead to overfitting.

Fig. 5 presents two test images overlaid with the corresponding pixel-wise bias, variance, and expected loss maps of the three different rectal cancer segmentation networks. Because most negative (noncancerous) regions show neither bias nor variance, we cropped the images according to the yellow box in column (d) for ease of visualization. Column (d) also illustrates the cropped MR image as well as the ground truth overlaid with the cropped MR image. The yellow arrows in columns (a) and (b) indicate distinct regions compared to the map below. For example, arrows on the variance map created by MRSN indicate the regions where major difference in variance between MRSN and MRSN-AUG occurs. The variance maps visualize the regional robustness of the rectal cancer segmentation networks. Moreover, combined with the bias maps, the variance maps create the expected loss map, thus visualizing how variance affects the expected loss.

Fig. 5 suggests that bias and variance tend to occur at the boundary of the rectal cancer regions, which may call for a loss function that weighs the border region more heavily than other regions, as suggested by Ronneberger *et al.* [14]. Otherwise, the general segmentation loss and the boundary segmentation loss can be treated as separate tasks and be merged by adding their respective losses, as suggested by Wang *et al* [33]. The yellow arrows suggest that both adding an additional rectum segmentation task and the augmentation method based on image resizing reduce the variance (or bias) of rectal cancer segmentation networks in the positive regions.

**TABLE 1.** Expected values of bias, variance, and expected zero-one loss within the positive and the entire region.

| | Positive | | | Total | | |
|---|---|---|---|---|---|---|
| | E[Bias] | E[Var] | E[L] | E[Bias] | E[Var] | E[L] |
| SRSN | 0.248 | 0.093 | 0.272 | 0.003 | 0.001 | 0.003 |
| | ± 0.232 | ± 0.065 | ± 0.225 | ± 0.004 | ± 0.001 | ± 0.004 |
| MRSN | 0.232 | **0.084** | **0.246** | 0.003 | 0.001 | 0.003 |
| | ± 0.232 | **± 0.054** | **± 0.214** | ± 0.004 | ± 0.001 | ± 0.004 |
| MRSN-AUG | 0.225 | **0.075** | 0.243 | 0.003 | **0.001** | 0.003 |
| | ± 0.227 | **± 0.050** | ± 0.220 | ± 0.004 | **± 0.001** | ± 0.004 |

All quantities are expressed as mean ± standard deviation. Any quantity shown in bold for MRSN or MRSN-AUG are significantly different ($p < 0.05$) from that of SRSN or MRSN, respectively.

Furthermore, the expected values over the entire image as well as over the positive regions, which were described in Section II D, were used to assess the efficacy of our two proposed methods in reducing the model variance of rectal cancer segmentation networks. Table 1 shows that a significant difference ($p < 0.05$, paired t-test) in model variance within the positive region was observed between SRSN and MRSN as well as between MRSN and MRSN-AUG. Adding the rectum segmentation task decreased the variance of rectal cancer segmentation by a factor of 0.90, whereas the augmentation method further lowered the variance by a factor of 0.89, on average. Moreover, the augmentation based on image resizing significantly reduced the variance of rectal cancer segmentation networks over the entire image. Neither approach increased the bias. Instead, both approaches decreased the bias, although not in a statistically significant amount.

In the future, the bias-variance decomposition using DSC as a loss function will be an interesting topic of investigation since DSC is a widely used metric in medical image analysis. Given that our main focus was to confirm that rectum information could improve rectal cancer segmentation, we left the design of an elaborate task-specific layer for both rectum and rectal cancer segmentation to future research. It has to be noted that our network has a limitation in that it neglects the information from neighboring image slices. The investigation of 3D segmentation network with stacked rectal MR images will be an interesting topic for future investigation. Although previous studies have also used 2D images in order to reduce the computational complexity, the 3D network can exploit information along the vertical axis [34, 35]. Both the segmentation performance and the bias-variance map of the 2D network can also be compared to those of the 3D network.

### B. PERFORMANCE EVALUATION WITH CROSS-VALIDATION

We asked whether our approaches to reducing model variance would show improvement also using conventional evaluation methods. The performance of SRSN, MRSN, and MRSN-AUG were compared using the method described in Section II E, and the results are reported in Table 2.

Significant differences in tumor DSC as well as in tumor sensitivity ($p < 0.05$, paired t-test) were observed between SRSN and MRSN. The augmentation method based on image resizing also improved the segmentation performance of MRSN in tumor DSC, rectum DSC, tumor sensitivity, and rectum specificity, with statistical significance.

Our approaches reduced the training duration as well. The training of the MRSN networks took less time than that of the rectal cancer and rectum SRSNs by an average factor of 0.96 and 0.81, respectively. The augmentation further reduced the MRSN training duration by a factor of 0.78 on average.

**TABLE 2.** Performance evaluation with 10-fold cross-validation using Dice Similarity Coefficient (DSC), sensitivity, and specificity.

| | DSC | | Sensitivity | | Specificity | |
|---|---|---|---|---|---|---|
| | Rectal Cancer | Rectum | Rectal Cancer | Rectum | Rectal Cancer | Rectum |
| SRSN | 0.723 | 0.940 | 0.743 | 0.938 | 0.999 | 0.999 |
| | ± 0.204 | ± 0.088 | ± 0.232 | ± 0.107 | ± 0.002 | ± 0.002 |
| MRSN | **0.732** | 0.938 | **0.755** | 0.936 | 0.999 | 0.999 |
| | **± 0.195** | ± 0.079 | **± 0.221** | ± 0.104 | ± 0.002 | ± 0.002 |
| MRSN-AUG | **0.742** | **0.943** | **0.765** | 0.939 | 0.999 | **0.999** |
| | **± 0.185** | **± 0.072** | **± 0.216** | ± 0.098 | ± 0.001 | **± 0.001** |

All quantities are expressed as mean ± standard deviation. Any quantity shown in bold for MRSN or MRSN-AUG are significantly different ($p < 0.05$) from that of SRSN or MRSN, respectively.

## IV. CONCLUSION

Deep learning has improved the state-of-the-art in various computer vision-related tasks, including image segmentation. Although most deep learning-based models were trained on large datasets, medical datasets are usually more limited in size [36-38]. In particular, annotated data for medical image segmentation are especially scarce due to the difficulty of manual delineation. Consequently, deep learning-based medical segmentation models risk suffering from model variance, which can cause overfitting. Methods able to reduce and evaluate the variance of segmentation models are thus important.

In this study, we suggested methods to measure and reduce the model variance of a rectal cancer segmentation model. First, we proposed a method for the pixel-wise bias-variance analysis of segmentation networks. This method can visualize the map of bias, variance, and expected loss, and also quantify their expected values within an arbitrary ROI. Second, we exploited the geometric correlation between the rectum and rectal cancer to reduce the model variance of the deep learning-based rectal cancer segmentation network. Lastly, we performed data augmentation by resizing mini-batches of images to further reduce the model variance. Such an approach was motivated by the common scale heterogeneity of medical imaging datasets.

To prove the efficacy of these two approaches in reducing variance without increasing bias, we tested the proposed pixel-wise bias-variance analysis method. Both approaches successfully reduced the model variance, especially within the positive region, without increasing the bias, and reduced the training duration as well. The efficacy of our approaches was also confirmed by using DSC via 10-fold cross-validation. Besides, our encoder-decoder segmentation network improves the computational efficiency of a previous study of rectal cancer segmentation as well [6]. Clinically, our network can effectively assist radiologists, because the demarcation of both rectum and rectal cancer is required for rectal cancer staging. By reducing the model variance, our approach will improve the accuracy of rectal cancer staging as well. Other cancer segmentation networks may be inspired by our approach to lowering the variance by exploiting the geometric correlation between cancer and the organ from which cancer grows. In our future research, we will develop a 3D segmentation network with stacked medical images. We will develop a 3D rectal cancer segmentation network and compare its performance with that of the 2D network.

## APPENDICES

### A. CONVOLUTION BLOCK STUDY

This section describes the details of our network. Our encoder-decoder network (Fig. 2) involves seven convolution blocks for which we have selected the best block among three candidates as illustrated in Fig. 6. Specifically, block3 adopts two consecutive residual connections whereas block1 and block2 are conventional VGG-style convolutions without a skip connection and a conventional residual block, respectively [11]. For both block2 and block3, we implemented a pre-activation policy [11].

We compared the segmentation performance of three different blocks on MRSN using the method described in Section II E. As described in Table 3, block3 scored the highest DSC for rectal cancer segmentation tasks and thus was selected as our convolution block. In addition, it scored the best also for the rectum segmentation task. Using block3,

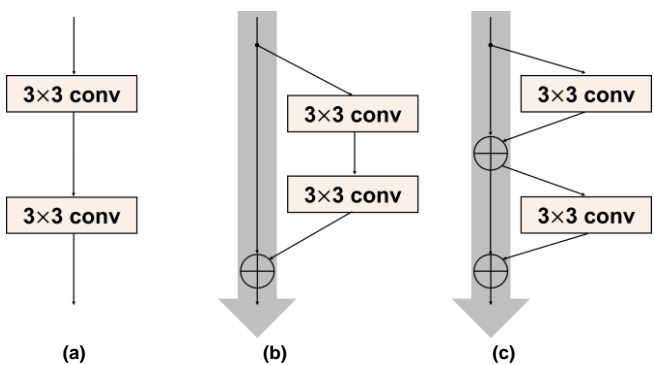

**FIGURE 6. Three candidates for our convolution blocks: (a) block1, (b) block2, and (c) block3.**

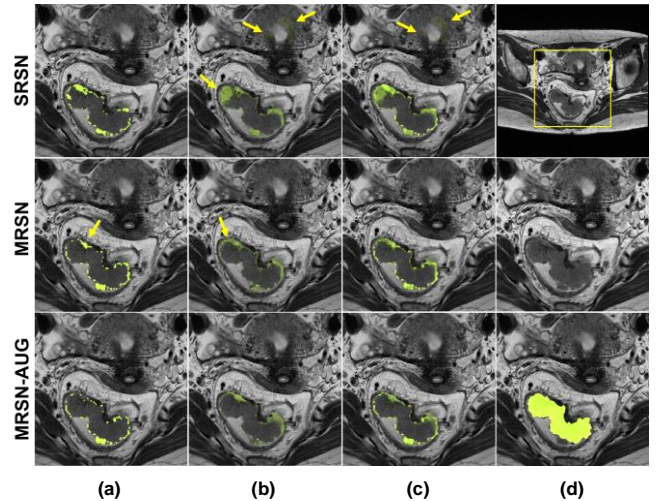

**(a)**   **(b)**   **(c)**   **(d)**

**FIGURE 7. A test image overlaid with pixel-wise bias, variance, and expected loss produced by three different rectal cancer segmentation networks, i.e., SRSN, MRSN, and MRSN-AUG. All settings are identical to the settings of Fig. 5.**

the network was trained faster than using the other two blocks (block1 was 1.46 times slower, and block2 was 1.04 times slower than block3).

**TABLE 3. Performance comparison between three different convolution blocks.**

|  | DSC | | Sensitivity | | Specificity | |
|---|---|---|---|---|---|---|
|  | Rectal Cancer | Rectum | Rectal Cancer | Rectum | Rectal Cancer | Rectum |
| Block 1 | 0.718 ± 0.201 | 0.935 ± 0.090 | 0.736 ± 0.225 | 0.932 ± 0.108 | 0.999 ± 0.002 | 0.999 ± 0.002 |
| Block 2 | 0.722 ± 0.204 | 0.936 ± 0.092 | 0.741 ± 0.230 | 0.937 ± 0.105 | 0.999 ± 0.002 | 0.999 ± 0.002 |
| Block 3 | 0.732 ± 0.195 | 0.938 ± 0.079 | 0.755 ± 0.221 | 0.936 ± 0.104 | 0.999 ± 0.002 | 0.999 ± 0.002 |

All quantities are expressed as mean ± standard deviation. and were obtained through 10-fold cross-validation. Block 1, 2, and 3 refer to the convolution blocks described in Fig. 6.

### B. BIAS-VARIANCE ANALYSIS WITHIN NON-CANCEROUS REGION

We focused on the positive regions to measure the expected values of bias, variance, and expected loss of the rectal cancer segmentation networks. However, bias and variance can also occur in negative areas, albeit rarely, and our method of exploiting the geometric correlation between rectum and rectal cancer can improve such problems. In Fig. 7, the SRSN model variance occurs at an organ with appearance similar to that of the rectum. Such variance at negative regions is removed by adding a rectum segmentation task to the network. Information about the rectum location can benefit the rectal cancer segmentation network since rectal cancer rarely exists outside of the rectum, especially in datasets, like our own, not containing tumors in the T4 group.

## ACKNOWLEDGMENT

We kindly thank Sun Ah Cho for her support in data acquisition.

## REFERENCES

[1] A. Bhandari, M. Woodhouse, and S. Gupta, "Colorectal cancer is a leading cause of cancer incidence and mortality among adults younger than 50 years in the USA: a SEER-based analysis with comparison to other young-onset cancers," *J. Investig. Med.,* vol. 65, no. 2, pp. 311-315, Feb. 2017.

[2] K.-W. Jung, Y.-J. Won, C.-M. Oh, H.-J. Kong, D. H. Lee, and K. H. Lee, "Prediction of cancer incidence and mortality in Korea, 2017," *Cancer Res. Treat.,* vol. 49, no. 2, pp. 306-312, Apr. 2017.

[3] M. Arnold, M. S. Sierra, M. Laversanne, I. Soerjomataram, A. Jemal, and F. Bray, "Global patterns and trends in colorectal cancer incidence and mortality," *Gut,* vol. 66, no. 4, pp. 683-691, Apr. 2017.

[4] M. K. Gospodarowicz, J. D. Brierley, and C. Wittekind, "Digestive System Tumours," in *TNM classification of malignant tumours,* 8th ed., Hoboken: John Wiley & Sons, 2017, pp. 84-88.

[5] S. H. Cho, Y. S. Cho, I. Y. Choi, H. I. Ha, J. Huh, B. Y. Hur, J. Hwang, H. Y. Jang, A. Y. Kim, and A. Y. Kim, "Essential Items for Structured Reporting of Rectal Cancer MRI: 2016 Consensus Recommendation from the Korean Society of Abdominal Radiology," *Korean. J. Radiol.,* vol. 18, no. 1, pp. 132-151, Jan.-Feb. 2017.

[6] S. Trebeschi, J. J. van Griethuysen, D. M. Lambregts, M. J. Lahaye, C. Parmer, F. C. Bakers, N. H. Peters, R. G. Beets-Tan, and H. J. Aerts, "Deep learning for fully-automated localization and segmentation of rectal cancer on multiparametric MR," *Sci. Rep.,* vol. 7, no. 1, p. 5301, Jul. 2017.

[7] Y. LeCun, Y. Bengio, and G. Hinton, "Deep learning," *Nature,* vol. 521, no. 7553, pp. 436-444, May 2015.

[8] A. Krizhevsky, I. Sutskever, and G. E. Hinton, "Imagenet classification with deep convolutional neural networks," in *Advances in Neural Information Processing Systems 25,* Red Hook, NY: Curran Associates, 2012, pp. 1097-1105.

[9] K. Simonyan and A. Zisserman, "Very deep convolutional networks for large-scale image recognition," *arXiv preprint arXiv:1409.1556,* 2014.

[10] C. Szegedy, W. Liu, Y. Jia, P. Sermanet, S. Reed, D. Anguelov, D. Erhan, V. Vanhoucke, and A. Rabinovich, "Going deeper with convolutions," in *Proceedings of 2015 IEEE Conference on Computer Vision and Pattern Recognition (CVPR),* Boston, MA, USA, 2015, pp. 1-9.

[11] K. He, X. Zhang, S. Ren, and J. Sun, "Identity mappings in deep residual networks," in *Computer Vision-ECCV 2016,* vol. 9908, B. Leibe, J. Matas, N. Sebe, and M. Welling, Eds. Cham: Springer International Publishing, 2016, pp. 630-645.

[12] G. Huang, Z. Liu, L. van der Maaten, and K. Q. Weinberger, "Densely connected convolutional networks," in *2017 IEEE Conference on Computer Vision and Pattern Recognition (CVPR),* Honolulu, HI, USA, 2017, pp. 2261-2269.

[13] G. Litjens, T. Kooi, B. E. Bejnordi, A. A. A. Setio, F. Ciompi, M. Ghafoorian, J. A. Van Der Laak, B. Van Ginneken, and C. I. Sánchez, "A survey on deep learning in medical image analysis," *Med. Image Anal.,* vol. 42, pp. 60-88, Dec. 2017.

[14] O. Ronneberger, P. Fischer, and T. Brox, "U-Net: Convolutional Networks for Biomedical Image Segmentation," in *Medical Image Computing and Computer-Assisted Intervention-MICCAI-2015,* vol. 9351, N. Navab, J. Hornegger, W. M. Wells, and A. F. Frangi, Eds. Cham: Springer International Publishing, 2015, pp. 234-241.

[15] J. Long, E. Shelhamer, and T. Darrell, "Fully convolutional networks for semantic segmentation," in *Proceedings of the IEEE conference on computer vision and pattern recognition(CVPR),* Boston, MA, USA, 2015, pp. 3431-3440.

[16] F. Milletari, N. Navab, and S.-A. Ahmadi, "V -Net: Fully Convolutional Neural Networks for Volumetric Medical Image Segmentation," in *Fourth International Conference on 3D Vision (3DV),* Stanford, CA, USA: IEEE, 2016, pp. 565-571.

[17] J. Hagerty, R. J. Stanley, and W. V. Stoecker, "Medical Image Processing in the Age of Deep Learning," in *Proceedings of the 12th International Joint Conference on Computer Vision, Imaging and Computer Graphics Theory and Applications (VISIGRAPP 2017),* Porto, Portugal, 2017, pp. 306-311.

[18] I. Goodfellow, Y. Bengio, A. Courville, and Y. Bengio, "Machine Learning Basics," in *Deep learning,* vol. 1, Cambridge, MA: MIT press, 2016, pp. 98-164.

[19] S. Ioffe and C. Szegedy, "Batch normalization: Accelerating deep network training by reducing internal covariate shift," *arXiv preprint arXiv:1502.03167,* 2015.

[20] I. J. Goodfellow, D. Warde-Farley, M. Mirza, A. Courville, and Y. Bengio, "Maxout networks," *arXiv preprint arXiv:1302.4389,* 2013.

[21] N. Srivastava, G. Hinton, A. Krizhevsky, I. Sutskever, and R. Salakhutdinov, "Dropout: a simple way to prevent neural networks from overfitting," *J. Mach. Learn. Res.,* vol. 15, no. 1, pp. 1929-1958, Jun. 2014.

[22] S. Ruder, "An overview of multi-task learning in deep neural networks," *arXiv preprint arXiv:1706.05098,* 2017.

[23] P. Domingos, "A unified bias-variance decomposition and its application," in *Proc. 17th International Conf. on Machine Learning,* Stanford, CA, 2000, pp. 231-238.

[24] M. Drozdzal, G. Chartrand, E. Vorontsov, M. Shakeri, L. Di Jorio, A. Tang, A. Romero, Y. Bengio, C. Pal,

and S. Kadoury, "Learning normalized inputs for iterative estimation in medical image segmentation," *Med. Image Anal.,* vol. 44, pp. 1-13, Feb. 2018.

[25] R. Srivaramangai, P. Hiremath, and A. S. Patil, "Preprocessing MRI images of colorectal cancer," *IJCSI,* vol. 14, no. 1, pp. 48–59, Jan. 2017.

[26] K. Zuiderveld, "Contrast limited adaptive histogram equalization," in *Graphics Gems IV*, Amsterdam, The Netherlands: Elsevier, 1994, pp. 474-485.

[27] B. B. Singh and S. Patel, "Efficient medical image enhancement using CLAHE enhancement and wavelet fusion," *IJCA,* vol. 167, no. 5, pp. 1-5, Jun. 2017.

[28] T. Dao, A. Gu, A. J. Ratner, V. Smith, C. De Sa, and C. Ré, "A kernel theory of modern data augmentation," *arXiv preprint arXiv:1803.06084,* 2018.

[29] T.-Y. Lin, P. Dollár, R. Girshick, K. He, B. Hariharan, and S. Belongie, "Feature pyramid networks for object detection," in *Proceedings of the IEEE conference on computer vision and pattern recognition(CVPR)*, Honolulu, HI, USA, 2017, pp. 2117-2125.

[30] K. He, X. Zhang, S. Ren, and J. Sun, "Delving deep into rectifiers: Surpassing human-level performance on imagenet classification," in *Proceedings of the IEEE International Conference on Computer Vision (ICCV)*, Santiago, Chile, 2015, pp. 1026-1034.

[31] X. Glorot and Y. Bengio, "Understanding the difficulty of training deep feedforward neural networks," in *Proceedings of the Thirteenth International Conference on Artificial Intelligence and Statistics*, Chia Laguna, Italy, 2010, pp. 249-256.

[32] D. P. Kingma and J. Ba, "Adam: A method for stochastic optimization," *arXiv preprint arXiv:1412.6980,* 2014.

[33] L. Wang, H. Zhen, X. Fang, S. Wan, W. Ding, and Y. Guo, "A unified two-parallel-branch deep neural network for joint gland contour and segmentation learning," *Future Generation Computer Systems,* vol. 100, pp. 316-324, Nov. 2019.

[34] Ö. Çiçek, A. Abdulkadir, S. S. Lienkamp, T. Brox, and O. Ronneberger, "3D U-Net: Learning Dense Volumetric Segmentation from Sparse Annotation," in *Medical Image Computing and Computer-Assisted Intervention – MICCAI 2016*, S. Ourselin, L. Joskowicz, M. R. Sabuncu, G. Unal, and W. Wells, Eds. Cham: Springer International Publishing, 2016, pp. 424-432.

[35] Y. Zhao, H. Li, S. Wan, A. Sekuboyina, X. Hu, G. Tetteh, M. Piraud, and B. Menze, "Knowledge-Aided Convolutional Neural Network for Small Organ Segmentation," *IEEE J. Biomed. Health Inform,* vol. 23, no. 4, pp. 1363-1373, Jan. 2019.

[36] M. Everingham, S. A. Eslami, L. Van Gool, C. K. Williams, J. Winn, and A. Zisserman, "The pascal visual object classes challenge: A retrospective," *Int. J. Comput. Vis.,* vol. 111, no. 1, pp. 98-136, Jan. 2015.

[37] T.-Y. Lin, M. Maire, S. Belongie, J. Hays, P. Perona, D. Ramanan, P. Dollár, and C. L. Zitnick, "Microsoft

COCO: common objects in context," in *Computer vision – ECCV 2014*, vol. 8693, D. Fleet, T. Pajdla, B. Schiele, and T. Tuytelaars, Eds. Cham: Springer International Publishing, 2014, pp. 740-755.

[38] J. Deng, W. Dong, R. Socher, L.-J. Li, K. Li, and L. Fei-Fei, "Imagenet: A large-scale hierarchical image database," in *2009 IEEE conference on computer vision and pattern recognition*, Miami, FLA, USA, 2009, pp. 248-255.

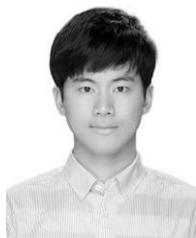

**JOOHYUNG LEE** was born in Seoul, South Korea in 1988. He received his B.S. degree in electrical engineering from the University of Texas at Dallas, Richardson, U.S., in 2012 and his M.S. degree in electrical engineering from Texas A&M University, College Station, U.S., in 2014. In 2016, he joined the Division of Convergence Technology of the National Cancer Center Research Institute in South Korea, where he is currently conducting research on the application of machine learning to medical data. He currently serves as a reviewer for the IEEE Journal of Biomedical and Health Informatics.

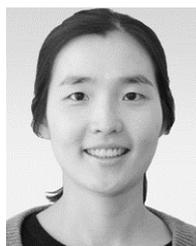

**JI EUN OH** received her M.S. and Ph.D. degrees in radiological science from Yonsei University, South Korea, in 2008 and 2013, respectively. She has been working as a researcher at the National Cancer Center Research Institute, South Korea. Currently, her research interests include multimodal x-ray imaging and radiomics, and especially computer-aided diagnosis and deep learning methods for precision medicine and its applications.

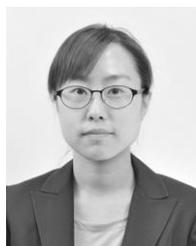

**MIN JU KIM** was born in Seoul, South Korea in 1977. She received her M.D. and M.S. degrees from Jungang University Medical School and Chungman National University Medical School, South Korea, in 2001 and 2015, respectively. She is an expert in the field of abdominal imaging, especially gastrointestinal radiology. She is a radiologist with 18 years of experience.

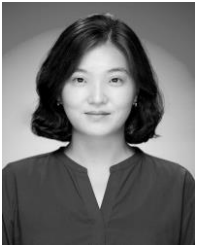

**BO YUN HUR** received her M.D. and M.S. degrees from Seoul National University Medical School, South Korea, in 2008 and 2014, respectively. She is an expert in the field of abdominal imaging, especially gastrointestinal radiology. She has extensive experience of rectal MRI reading, amounting to more than 300 studies per year. She is a radiologist with 10 years of experience.

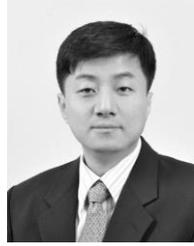

**DAE KYUNG SOHN** received his M.D. from the Seoul National University Medical School, South Korea, in 1997 and his Ph.D. degree from Chungbuk National University, South Korea in 2011. He is a surgeon at the Center for Colorectal Cancer of the National Cancer Center Korea. He is also the Director of the Healthcare Platform Center and the chief of the Department of Innovation Technology of the National Cancer Center Korea. He is an expert in the fields of therapeutic endoscopy, natural orifice surgery, and medical engineering.